# Prediction Interval Construction Method for Electricity Prices

Xin Lu, The University of Sydney

*Abstract*-- **Accurate prediction of electricity prices plays an essential role in the electricity market. To reflect the uncertainty of electricity prices, price intervals are predicted. This paper proposes a novel prediction interval construction method. A conditional generative adversarial network is first presented to generate electricity price scenarios, with which the prediction intervals can be constructed. Then, different generated scenarios are stacked to obtain the probability densities, which can be applied to accurately reflect the uncertainty of electricity prices. Furthermore, a reinforced prediction mechanism based on the volatility level of weather factors is introduced to address the spikes or volatile prices. A case study is conducted to verify the effectiveness of the proposed novel prediction interval construction method. The method can also provide the probability density of each price scenario within the prediction interval and has the superiority to address the volatile prices and price spikes with a reinforced prediction mechanism.**

*Index Terms*-- **Generative Adversarial Networks (GAN); Electricity Price; Prediction Intervals; Spike Prediction; Weather Factors**

## I. INTRODUCTION

ELECTRICITY price prediction has attracted considerable attentions [1-3] because an accurate prediction of electricity prices plays an essential role [4-6] in supporting the behaviors of market participants, such as profit optimization, risk control, and stable operation. Electricity prices are usually determined by several dependencies and are generally regarded as anti-persistent data. Based on a study of the marketability of variable wind power generation and solar power generation in the electricity market, Zipp [7] calculated the merit-order effect and indicated a systematic decrease in electricity market revenues for wind and solar plants. Mosquera et al. [8] studied the impact of weather factors on electricity prices, including temperature, wind speed, precipitation, and irradiance. These weather factors can affect the demand for electricity and wind or solar power generation, resulting in the fluctuation of electricity prices and bringing significant challenges to electricity price prediction.

Electricity price prediction can be classified into point prediction and interval prediction. The point prediction is usually a deterministic prediction, where the predicted points are obtained by fitting methods with the root mean square error (RMSE) and mean absolute error (MAE) as the minimization objects between the predicted data and the actual data [9, 10]. However, uncertainties are inevitable [11], usually due to incomplete data or unanticipated events. In recognition of the inherent limitations of point prediction models, interval prediction is developed to describe the uncertainties and compensate for the low accuracy of point prediction.

Extensive research has been reported in the literature on the construction of interval prediction models. Statistical methods are often used for interval prediction. In [12], an autoregressive integrated moving average (ARIMA) approach was employed to estimate the confidence interval of electricity prices by assuming that the residuals follow uniform distributions. Machine learning (ML) methods are also widely employed in interval prediction due to their outstanding performance in handling nonlinear problems. The mean-variance estimation (MVE) technique combined with artificial neural network (ANN) was applied for wind power predictions in [13]. This method assumes that the distribution of errors should be an additive Gaussian distribution with non-constant variance. Khosravi et al. [14] proposed a lower upper bound estimation (LUBE) method for prediction intervals, directly giving the upper bound and lower bound with ANN. A multi-objective artificial bee colony (MOABC) modified wavelet neural network (WNN) was proposed by Shen et al. [15] to achieve good interval prediction performance. A Pareto optimal price LUBE model was established in [16], combining non-dominated sorting genetic algorithm II (NSGA-II) and extreme learning machine (ELM) to give intervals directly. The LUBE-based method can predict well the intervals for smooth time series but is handicapped for high volatile scenarios [17]. These interval prediction methods, however, can only give the upper and lower bounds of intervals. They cannot estimate the probability densities inside the intervals. The internal probability density is of great significance in accurately reflecting the uncertainty of electricity prices and for providing a data foundation for price optimization or risk management.

A generative adversarial network (GAN) [18-20] is a new type of deep learning model that generally consists of a generator and a discriminator. Chen et al. [21] proposed a GAN-based wind forecasting model. However, a good GAN model for time-series data should preserve the temporal dynamics, and the generated scenarios should respect the original distribution and relationship between variables across the whole period. Yoon [22] introduced the latent space to GAN and proposed a time series GAN (TSGAN) framework for synthesizing realistic scenarios, which integrates the versatility of unsupervised training with the control of supervised training, and has a perfect predictive ability. GAN was originally invented to generate a large number of different real-like images [23], and this same property can be used to generate a large number of real-like time series to simulate the actual time series probability density distribution [24]. Currently, there is no research on this.

This paper proposes a novel scenario generation based prediction interval construction method, considering different volatility levels of weather factors. The major novel contributions of this work are as the following:
1) A novel CTSGAN-based interval prediction method is proposed, which can capture the distribution of each period and preserve temporal variations and generate realistic price scenarios.
2) With realistic price scenarios, the prediction intervals can be constructed, and the predictive probability density of each scenario can be estimated, which reflects uncertainty accurately.
3) The uncertainty of electricity supply and demand can be estimated according to the volatility level of weather factors. The CTSGAN model with a reinforced prediction mechanism can generate pattern-diversity prediction intervals, which can be used to address spikes in electricity prices.

The rest of this paper is organized as follows: Section II introduces the new evaluation indicators of prediction intervals. Section III investigates the effect of weather factors on electricity prices. Section IV presents the steps of intervals prediction for normal trend prices and volatile prices. Section V conducts the case studies, followed by the conclusion and some discussions on future research directions.

## II. EVALUATION INDICATORS OF PREDICTION INTERVALS

Reliability and sharpness are two indispensable indicators of prediction intervals [11]. Khosravi et al. [14] introduced the empirical coverage probability of all samples (ECPAS) and the empirical average width of all prediction intervals (EAWAPI) to evaluate the reliability and sharpness of prediction intervals. However, Neyman [25] pointed out that confidence level is usually calculated by repeated sampling. It is essential to note that ECPAS does not equate to a confidence level, the target of ECPAS is all the test samples (several electricity prices) and the target of confidence level is just one sample (one electricity price), which means that ECPAS cannot guild the LUBE method to construct reasonable intervals. To properly evaluate the prediction methods, we redefined the interval prediction indicators. According to the definition of the confidence interval [25], the definition of confidence level for ECPAS and EAWAPI can be given as follows.

**Definition 1** (Confidence level for a Y% ECPAS) *An X% confidence level for a Y% ECPAS is an ECPAS calculated by prediction intervals generated by a procedure that in repeated sampling has an X% probability of being larger than or equal to Y%, for all possible ECPAS.*

**Definition 2** (Confidence level for an EAWAPI Z) *An X% confidence level for EAWAPI Z is an EAWAPI calculated by prediction intervals generated by a procedure that in repeated sampling has an X% probability of being smaller than or equal to Z, for all possible EAWAPI.*

The definitions of ECPAS $\delta^s$ and confidence level for ECPAS $\phi$ are given as follows:

$$\delta^s = \frac{1}{T}\sum_{t=1}^{T} c_t, \quad \begin{cases} c_t = 1, & \text{if } \theta_t \in [L_t^s, U_t^s] \\ c_t = 0, & \text{if } \theta_t \notin [L_t^s, U_t^s] \end{cases} \quad (1)$$

$$\phi = \frac{1}{S}\sum_{s=1}^{S} b^s, \quad \begin{cases} b^s = 1, & \text{if } \delta^s \geq \delta' \\ b^s = 0, & \text{if } \delta^s < \delta' \end{cases} \quad (2)$$

where $T$ represents the sample size, $\theta_t$ the $t^{\text{th}}$ sample, $S$ the total number of repeated prediction processes, and $s$ the $s^{\text{th}}$ prediction process. $c_t$ is an indicator that is equal to 1 if $s^{\text{th}}$ prediction interval $[L_t^s, U_t^s]$ covers $\theta_t$, and to 0 otherwise. $b^s$ denotes another indicator, it equals 1 if $\delta^s$ no less than a certain value of ECPAS, $\delta'$, and otherwise 0.

The definitions of EAWAPI $\xi^s$ and confidence level for EAWAPI $\varphi$ are given as follows.

$$\xi^s = \frac{1}{T}\sum_{t=1}^{T}\left(U_t^s - L_t^s\right) \quad (3)$$

$$\varphi = \frac{1}{S}\sum_{s=1}^{S} a^s, \quad \begin{cases} a^s = 1, & \text{if } \xi^s < \xi' \\ a^s = 0, & \text{if } \xi^s \geq \xi' \end{cases} \quad (4)$$

where $a^s$ denotes an indicator. It equals 1 if $\xi^s$ is no less than a certain EAWAPI with $\xi'$, and otherwise 0.

Fig. 1 illustrates a simple example of the difference between a Y% ECPAS and a confidence level for an ECPAS Y%. Fig.1 (a) shows the empirical coverage probability of all samples. There are 20 actual samples, and the prediction intervals do not cover two samples; thus, the ECPAS is 90%. This prediction process is repeated 50 times, and the ECPAS is different each time, as shown in Fig. 1 (b). The ECPAS being larger than or equal to 90% (above the dotted red line) occurs 40 times out of 50, known as a 90% confidence level. This situation can be said that the ECPAS will be larger than or equal to 90% with a confidence level of 90%. By changing the ECPAS target value $\delta'$, another conclusion can be obtained that a 95% confidence level for ECPAS will be equal to or larger than 85% (above the solid red line).

## III. EFFECT OF WEATHER FACTORS ON DAY-AHEAD ELECTRICITY PRICES

### A. Related factors of electricity prices

The supply and demand for electricity determine the electricity prices [26]. The electricity supply consists of two main components, conventional power plants and renewable energy generation. Factors affecting the cost of conventional power plants include the prices of coal and gas, and factors affecting the power generation from renewable energy resources include wind and solar conditions. The electricity demand is due to several factors, including local population, economy, and industrial development for long-term consideration, and for the short term, the temperature and the day of the week are the most critical factors.

### B. Impact of weather volatility on electricity prices

According to the Australian Energy Regulator (AER) reports [27], most electricity price spikes are due to high temperatures and sudden weather changes. Typically, mid-day is the time of day when electricity demand is highest. However, due to the widespread use of rooftop photovoltaics in Australia, the daily electricity demand peak has shifted to the afternoon [28, 29]. Even if the weather fluctuates in the demand off-peak period, the regular electricity supply can be realized through the cooperation of multiple energy sources to avoid electricity price spikes. During the peak period of electricity demand, it will further increase the demand if there is a sudden high temperature. If renewable energy generation is reduced due to weather factors, it will lead to fluctuations in electricity prices and even spikes. In addition, there is an increase in rebidding behavior in the electricity market due to weather changes and difficulties in accurately estimating the wind and solar power output, which can also lead to fluctuations in electricity prices

[27]. Fig. 2 shows a daily pattern of the number of spikes from 2016 to 2020. Interestingly, extreme high electricity price spikes (≥ A$350/MWh) only occur from 12:00 to 19:00.

Inspired by [8], we collected the electricity price, temperature, and wind speed data between 12:00 and 19:00 and shortwave irradiance data between 12:00 and 17:00 (low irradiance after 17:00) and calculated the daily variance of each type of data for the years of 2016 to 2020. Table I lists the Pearson correlation analysis between the variance of each weather factor and the variance of the electricity price.

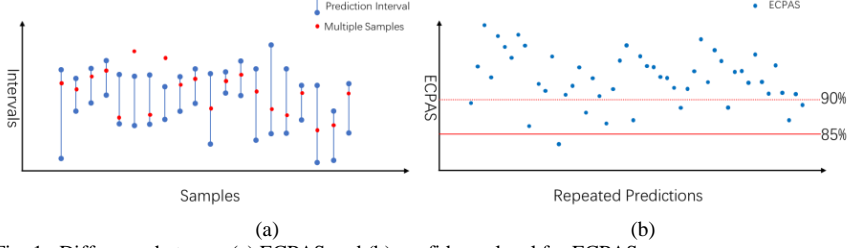

Fig. 1. Difference between (a) ECPAS and (b) confidence level for ECPAS

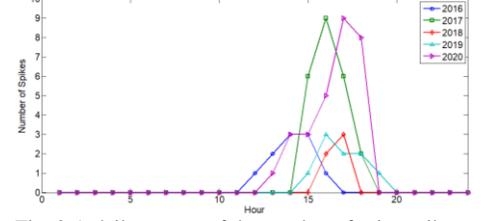

Fig. 2 A daily pattern of the number of price spikes (no less than A$350/MWh) from 2016 to 2020

As shown in Table I, the variance of electricity prices and that of weather factors show medium linear correlations, in the order of correlation being temperature, irradiance, and wind speed, which means that volatility in temperature, irradiance, and wind speed can directly result in volatility in the price of electricity. Table II tabulates the volatility level based on the statistical results of each weather factor; 60% of the volatility of weather factors are within the normal range of variation, the remaining 40% volatility is divided into three levels: high, medium, and low with 25%, 10%, and 5%, respectively.

TABLE I
PEARSON CORRELATION ANALYSIS OF PRICE AND WEATHER FACTORS

|  | R-value | P-value |
| --- | --- | --- |
| Price (Var.) and Temperature (Var.) | 0.6382 | <0.0001 |
| Price (Var.) and Shortwave Irradiance (Var.) | 0.5911 | <0.0001 |
| Price (Var.) and Wind Speed (Var.) | 0.5239 | <0.0001 |

TABLE II
VOLATILITY LEVEL OF WEATHER FACTORS

|  | Normal (<60%) | Level Low (60%~85%) | Level Medium (85%~95%) | Level High (>95%) |
| --- | --- | --- | --- | --- |
| Temperature | [0,0.0019] | [0.0019,0.0030] | [0.0030,0.0058] | [0.0058,1] |
| Irradiance | [0,0.0246] | [0.0246,0.0419] | [0.0419,0.0622] | [0.0622,1] |
| Wind Speed | [0,0.0052] | [0.0052,0.0079] | [0.0079,0.0173] | [0.0173,1] |

## IV. PREDICTION STEPS AND METHODOLOGY

### A. Scenario generation-based interval prediction

The conventional interval prediction methods are generally divided into two kinds. One is to estimate the mean and standard deviation separately and combine them to build prediction intervals, e.g., the MVE method. The other is to directly give the upper and lower bounds of the intervals, such as LUBE and Bootstrap methods. If all possible scenarios can be predicted, these scenarios can be combined to obtain an interval. This paper proposes a novel interval prediction method based on scenario generation.

### B. Interval prediction steps

Fig. 3 illustrates a flowchart of the interval prediction process, which consists of four main components: (a) training and optimization of CTSGAN, (b) normal price prediction, (c) reinforced prediction, which depends on the volatility of weather factors, and (d) combining the prediction scenarios into prediction intervals.

1) CTSGAN weight optimization

The Australian electricity prices and demands can be downloaded from the Australian Energy Market Operator (AEMO) website. To better fit the prediction model [30], the electricity prices are limited to a certain range [A$0/MWh, A$500/MWh], and the original data are normalized by

$$p' = \frac{p - p_{\min}}{p_{\max} - p_{\min}} \quad (5)$$

where $p_{\min}$ and $p_{\max}$ are the minimal and maximal prices in the dataset, and $p$ and $p'$ the original and normalized values, respectively.

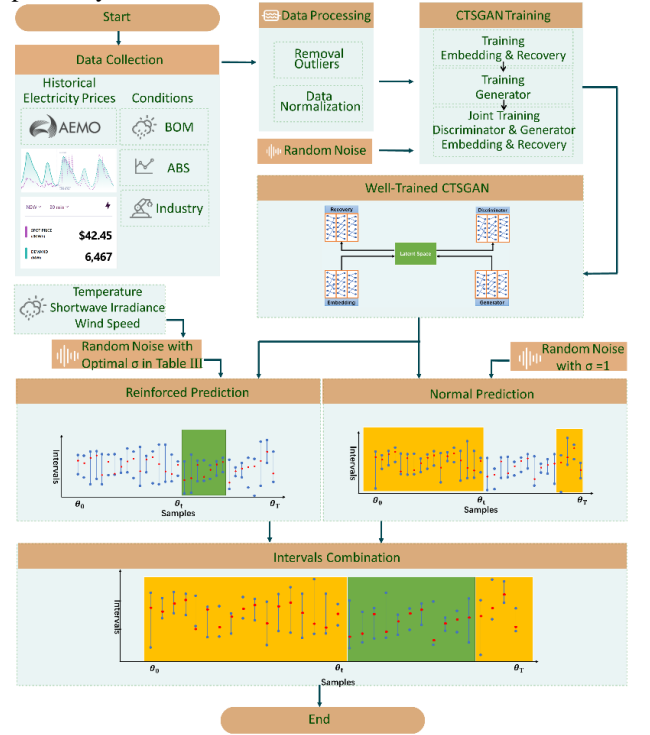

Fig. 3 Flowchart of interval prediction

The inputs of the CTSGAN are the pre-processing data, random noises, and conditions. The training process can be completed in three steps: the first optimization objectives are the weights of the embedding and recovery networks; Then, the optimization objectives are the weights of the generator; The final four networks joint training.

2) Prediction of normal scenarios

For the normal trend electricity prices, the inputs are conditions and random noises, which satisfy the standard Gaussian distribution, and different scenarios for the next 24 hours can be obtained with well-trained CTSGAN model

3) Reinforced prediction of volatile scenarios

By collecting the weather forecast information and calculating the volatility level of each weather factor, the volatility of electricity prices in the afternoon time of the next day can be estimated. Conditions and selected distribution random noise input can be introduced to generate the pattern-diversity scenarios.

4) Interval combination and prediction

According to the step 3) and 4), the price scenarios for the next 24 hours can be generated, including the normal trend scenarios and volatile scenarios. Then, the predictive probability density of each half-hour price can be obtained by stacking the different scenarios, and prediction intervals can be obtained by combining different scenarios.

### C. CTSGAN

The novel interval prediction method is based on scenario generation, which can generate a sufficiently large number of scenarios. The conventional neural networks cannot generate a large number of different scenarios because the network parameters are fixed after the training is completed, and there is a one-to-one mapping relationship between the input and output. GAN is introduced here to generate different scenarios, in which the introduction of random noises as the input can diversify the generated scenarios. Moreover, the authenticity of the generated scenarios is also crucial for interval construction. As the conventional GAN is an unsupervised learning algorithm, the conventional GAN needs to be modified to a supervised algorithm to make the generated scenarios realistic. By combining the TSGAN and conditions, CTSGAN is first proposed in this paper to generate a large number of realistic and diversity-rich scenarios as the basis for the construction of intervals.

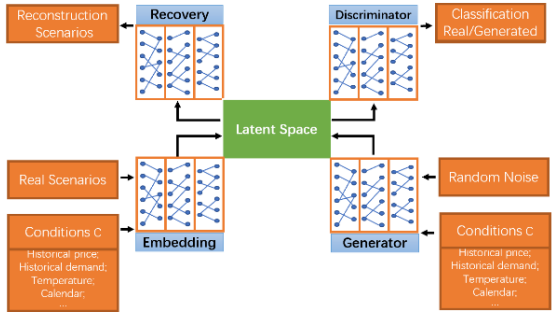

Fig. 4 Structure of the CTSGAN model

The GAN model is mainly composed of two neural networks, i.e., the generator and the discriminator. The responsibility of the generator is to generate a massive amount of realistic scenarios with different random noise inputs $z$ under a well-defined noise distribution (e.g., Gaussian distribution) until the discriminator cannot distinguish them from the real scenarios $x$. Denote the generator neural network and discriminator neural network as $G(z;\theta^{(G)})$ and $D(x;\theta^{(D)})$, and the weights of neural networks in the generator and discriminator as $\theta^{(G)}$ and $\theta^{(D)}$, respectively. Yoon [22, 31] proposed a TSGAN framework for synthesizing realistic scenarios, which integrates the versatility of unsupervised training with the control of supervised training. By combining the TSGAN and conditions, the CTSGAN is proposed, and the structure of CTSGAN is shown in Fig.4.

### D. Reinforced prediction of volatile scenarios

In the CTSGAN training and prediction process, it is generally assumed that the random noise $z$ follows a Gaussian distribution $\mathcal{N}(\mu,\sigma)$, and the standard Gaussian distribution, $\mathcal{N}(0,1)$, is commonly used. However, in the prediction process, $z \sim \mathcal{N}(0,1)$ will lead to poor predictions for some low-probability scenarios, like spikes or volatile scenarios.

A novel reinforced prediction mechanism in terms of volatility level of weather factors is first proposed and used in this paper to solve the problem of volatile scenarios. After CTSGAN is well-trained, the random noises satisfying different Gaussian distributions with different standard deviations are selected as the inputs, and then the reinforced CTSGAN (RCTSGAN) can generate more diverse scenarios. As shown in Fig. 5, the normal trend electricity prices (yellow part) can be predicted with random noise $z \sim \mathcal{N}(0,1)$, and the volatile scenarios (green part) can be predicted with random noise $z \sim \mathcal{N}(0,\sigma)$ in terms of volatility level of weather factors. The combined prediction intervals could be rich in diversity and contain more volatile scenarios and spikes. Table III lists the selection of $\sigma$ for random noise. Assume that the variances of the predicted temperature, irradiance, and wind speed are 0.004, 0.07, and 0.02, respectively. The standard deviation $\sigma$ of the random noise should be selected as 2.667 (1+1+0.667) according to Tables II and III.

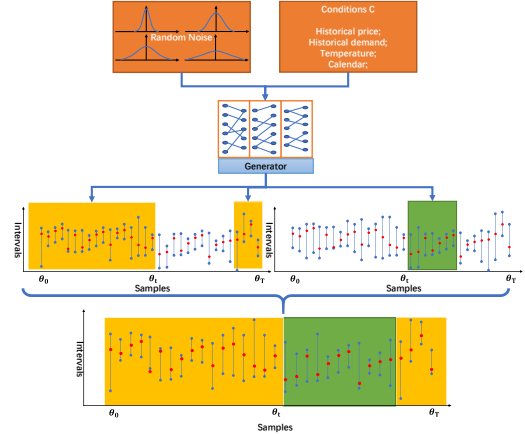

Fig. 5 Volatile scenarios reinforced prediction

Table III
SELECTION TABLE FOR $\sigma$ OF RANDOM NOISE

|  | Normal | Level Low | Level Medium | Level High |
| --- | --- | --- | --- | --- |
| Temperature | 0 | +0.333 | +0.667 | +1 |
| Irradiance | 0 | +0.333 | +0.667 | +1 |
| Wind Speed | 0 | +0.333 | +0.667 | +1 |

## V. CASE STUDY AND RESULT ANALYSIS

### A. Implementation details

For simplicity of the model, the long short term memory (LSTM) networks [31] are chosen as the core neural network of embedding function, recovery function, generator, and discriminator in CTSGAN and RCTSGAN. The number of hidden layer units in LSTMs is set to 100, and the number of iterations is set to 10000 for each training step. The other parameters for the proposed CTSGAN and RCTSGAN are set as follows: the batch size is 7; the number for each training iteration is 10,000; the clipping parameter is 0.5; the learning rate is 0.02; the dimensionality of latent space is 100. The parameters for the TSGAN are set as the same as those of the proposed CTSGAN and RCTSGAN. The artificial bee colony (ABC) optimization is employed to modify the weights of WNN in the LUBE method, and the parameters are set the same as those in [15].

The condition set contains many related factors, such as the half-hourly historical electricity price and demand, gas and coal spot prices, day type (day of the week and month of the year), heating degree day (HDD) and cooling degree day (CDD), and forecast weather factors, which includes shortwave irradiance, wind speed and temperature.

*B. Probability density of prediction intervals*

The conventional interval prediction methods, such as LUBE-based methods, only give the upper and lower bounds of prediction intervals but cannot give the probability density of each electricity price in the interval. Our proposed CTSGAN and RCTSGAN methods can generate prediction intervals and probability densities by combining the pattern-diversity scenarios. The visualized prediction scenarios are shown in Fig. 8(a), where the x, y, and z axes are the prediction time scale, the normalized prediction electricity price, and the probability density for each different price, respectively. The colors represent the probability densities of the occurrence of these scenarios. The x-axis [0,10] in Fig. 8(a) is partially enlarged, as shown in Fig. 8(b), which presents the detailed probability density of each scenario. The probabilities vary from time to time for different scenarios, with the highest probabilities ranging from 0.15 to 0.2. Fig. 8(c) presents the left view of Fig. 8(b). As shown, the probability density cross-section of each scenario for different times approximates a Gaussian distribution curve. This indicates that the pattern-diversity scenarios generated by CTSGAN or RCTSGAN follow approximately the normal distribution, consistent with the input random noise distribution. Fig. 8(d) is a top view of Fig. 8(a). Fig. 8(d) shows how scenarios with different probability densities can be combined into prediction intervals. Similar to Fig. 8(c), the scenarios with greater probabilities appear at the center of the prediction intervals, while scenarios of low probabilities stay near the upper and lower boundaries.

*C. Comparative simulation of prediction intervals with three different methods*

In Sections *V.B* and *V.C*, the price scenarios generated by CTSGAN are more realistic than those generated by TSGAN. In this section, the RCTSGAN, basic CTSGAN method, and LUBE method are applied to predict price intervals.

1) Prediction interval for volatile price scenarios

Three typical consecutive 5-day volatile electricity prices are chosen for analysis from the summer of 2017 (Fig. 9 (a), (b), and (c)), the summer of 2018 (Fig. 9 (d), (e), and (f)), and the summer of 2019 (Fig. 9 (g), (h) and (i)). The prediction intervals for the volatile price scenarios are shown in Fig. 9, where Figs. 9 (a), (d), and (g) are predicted by the RCTSGAN method, Figs. 9 (b), (e), and (h) by the basic CTSGAN method, and Fig. 9 (c), (f), and (i) by LUBE method with ECPAS of 90%, respectively.

As shown, the boundaries of the prediction intervals of the RCTSGAN and CTSGAN methods are not as smooth as the results of the LUBE method. This indicates that the two CTSGAN-based methods can retain more features during the training process, while the LUBE method fails to learn enough relevant features. Although the actual electricity price fluctuates wildly in Fig. 9 (c), the lower boundary of the LUBE-based prediction interval fluctuates slightly around 0.1. In addition, when the normalized electricity price exceeds 0.5 (A\$250/MWh), there is a higher probability that the LUBE prediction interval cannot contain the actual electricity price.

Figs. 9(a), (b), and (c) compare the prediction intervals for extremely high spikes of the successive five days. The spikes of the first and third days are about 0.7 (A\$350/MWh), and the spikes of the last two days are about 0.9 (A\$450/MWh). The prediction intervals obtained by the proposed RCTSGAN contain all extremely high spikes. However, the intervals given by both the basic CTSGAN and LUBE methods do not contain these spikes. For small spikes, all the three interval prediction methods show good performance and contain this spike.

Both RCTSGAN and CTSGAN perform well for predicting electricity prices during the daily off-spike period from 0:00 to 12:00 and from 19:00 to 24:00. However, the CTSGAN model is less effective in predicting sudden changes in spikes due to the volatility of weather factors. Overall, by evaluating the level of electricity price volatility from the forecasted weather conditions, the RCTSGAN model can choose optimal random noise input with $\sigma$ and introduce the reinforced prediction mechanism, enabling better prediction of electricity price intervals.

2) Prediction interval for normal price scenarios

For normal price scenarios, the CTSGAN method rarely triggers the spike-reinforced prediction mechanism. At this time, the RCTSGAN and CTSGAN methods give the same prediction intervals. Fig. 10 shows the prediction intervals with the CTSGAN method and LUBE method. It can be seen that the prediction intervals of both methods can include most of the electricity prices. However, the interval given by LUBE is significantly more expansive than that of CTSGAN.

3) Prediction intervals for four seasons from 2018 to 2020

The electricity prices of four seasons from 2018 to 2020 are chosen to verify the superiority of the proposed RCTSGAN and CTSGAN methods in comparison with the LUBE method. In the LUBE method, the ECPAS target is set to 95%, 90%, and 85%, respectively. Unlike the RCTSGAN or CTSGAN method, the WNN in LUBE cannot generate multiple prediction intervals. Therefore, it is necessary to repeat the training ten times to obtain WNNs that can generate ten predictions [32]. According to Definitions 1 and 2, the actual ECPAS, $\delta'$, and EWAPI, $\xi'$, with a 90% confidence level can be obtained, as listed in Table IV.

As can be seen from Table IV, some of the ECPAS obtained by the LUBE method can reach the intended target, like the 95% target for winter prices, while some cannot reach the target, such as the summer of 2019-20, the final predicted ECPAS is only 0.8471, even if LUBE sets the training target to 95%. This result may be explained that the ECPAS obtained by the LUBE method is related to the volatility of the predicted electricity prices themselves, and the prediction performance is more unsatisfactory when the electricity prices are with high levels of volatility. However, the RCTSGAN and CTSGAN methods predict the intervals by generating scenarios with as much diversity as possible. The RCTSGAN and CTSGAN methods are more suitable for electricity price prediction with ECPAS of 0.9678 (RCTSGAN) and 0.8916 (CTSGAN). As shown, the

TABLE IV
ECPAS AND EAWAPI OF PREDICTION INTERVALS (CONFIDENCE LEVEL = 90%)

| Year | Season | RCTSGAN | | CTSGAN | | LUBE (95%) | | LUBE (90%) | | LUBE (85%) | |
|---|---|---|---|---|---|---|---|---|---|---|---|
| | | ECPAS | EAWAPI | ECPAS | EAWAPI | ECPAS | EAWAPI | ECPAS | EAWAPI | ECPAS | EAWAPI |
| 2018 ~ 2019 | Spr. | **0.9589** | 0.2131 | 0.9393 | **0.2018** | 0.9364 | 0.2417 | 0.8712 | 0.2290 | 0.8091 | 0.2121 |
| | Sum. | **0.9363** | 0.2368 | 0.8850 | **0.2344** | 0.8711 | 0.2651 | 0.8535 | 0.2511 | 0.7810 | 0.2583 |
| | Fal. | **0.9572** | 0.2103 | 0.9389 | **0.1946** | 0.9466 | 0.2399 | 0.9034 | 0.2331 | 0.8418 | 0.2118 |
| | Win. | **0.9691** | 0.1934 | 0.9577 | **0.1863** | 0.9519 | 0.2521 | 0.9026 | 0.2529 | 0.8373 | 0.2463 |
| 2019 ~ 2020 | Spr. | **0.9345** | 0.2209 | 0.9242 | **0.1923** | 0.9118 | 0.2323 | 0.8742 | 0.2301 | 0.8645 | 0.2190 |
| | Sum. | **0.9678** | 0.2193 | 0.8916 | **0.2002** | 0.8474 | 0.2451 | 0.8329 | 0.2393 | 0.8144 | 0.2204 |
| | Fal. | **0.9717** | 0.2155 | 0.9378 | **0.2116** | 0.9381 | 0.2477 | 0.9155 | 0.2346 | 0.8521 | 0.2397 |
| | Win. | **0.9699** | 0.1964 | 0.9643 | **0.1821** | 0.9506 | 0.2634 | 0.9023 | 0.2448 | 0.8437 | 0.2216 |
| Average | | **0.9581** | 0.21321 | 0.9298 | **0.2004** | 0.9192 | 0.2484 | 0.8819 | 0.2393 | 0.8305 | 0.2286 |

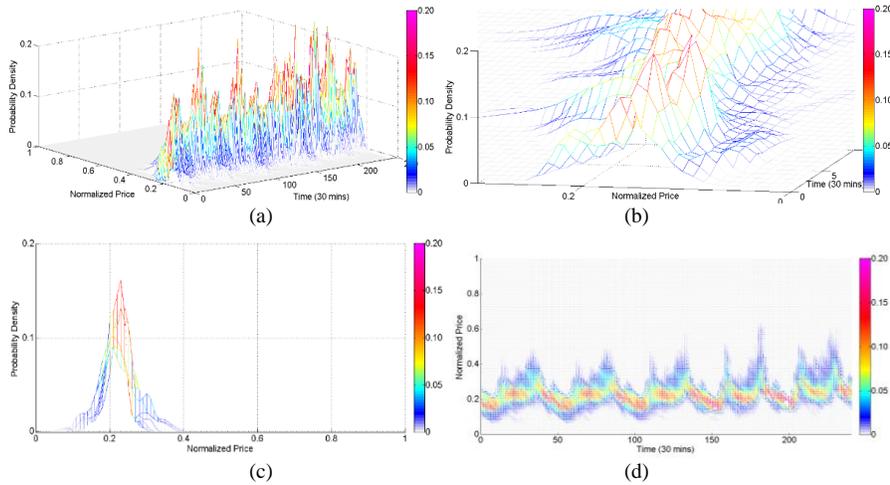

Fig. 8 Probability density of prediction scenarios, where (a) Overall view, (b) Enlarged partial view, (c) Left view (d) Top view

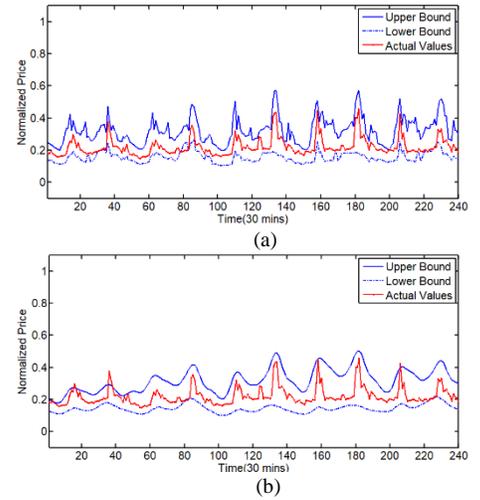

Fig. 10 Prediction interval for normal electricity price by (a) RCTSGAN/CTSGAN and (b) LUBE

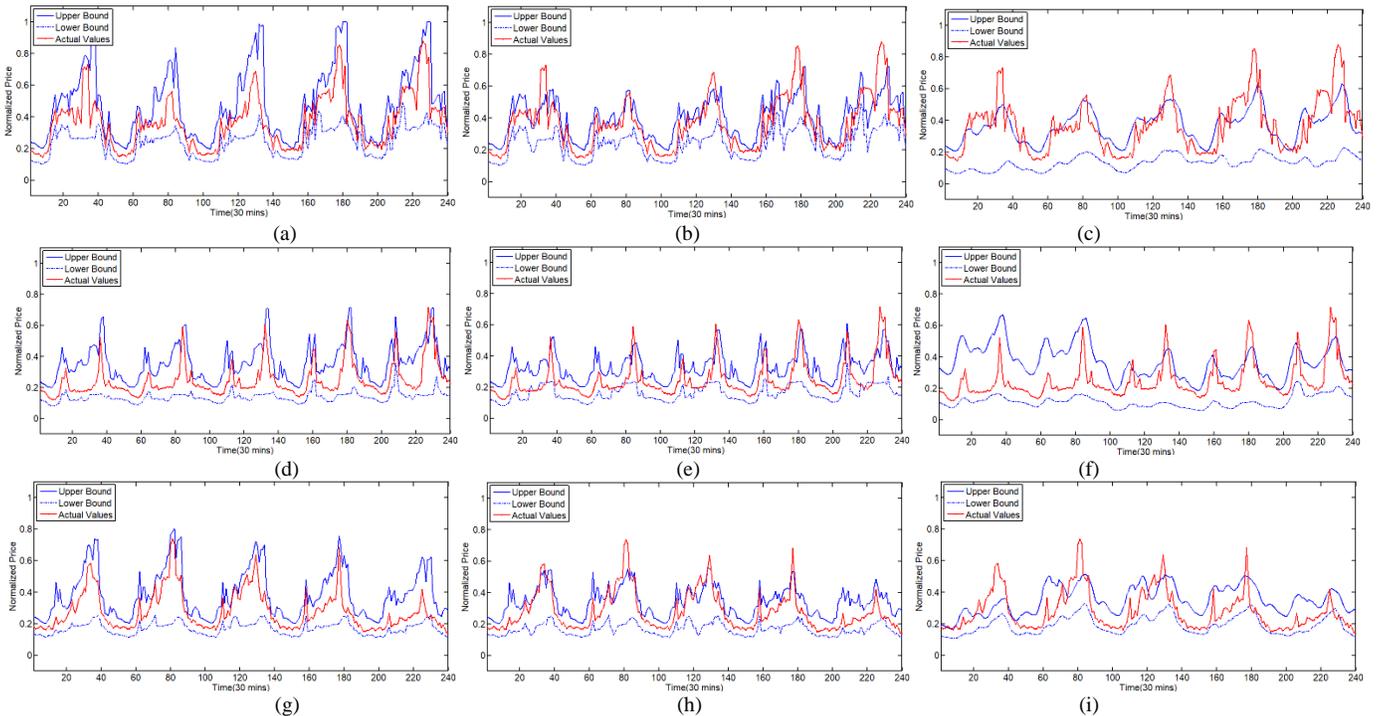

Fig. 9 Prediction intervals of volatile electricity price by three methods for three typical volatile prices, (a) (d) (g) RCTSGAN, (b) (e) (h) CTSGAN and (c) (f) (i) LUBE (ECPAS=90%)

scenario generation-based interval prediction methods significantly outperform the LUBE method.

In addition, the proposed RCTSGAN method gives accurate results for the prediction intervals with the highest average ECPAS of 0.9589 with a 90% confidence level and the second narrowest interval with an average EAWAPI of 0.21321. The basic CTSGAN method also shows good prediction performance with ECPAS of 0.9298 and EAWAPI of 0.2004. Introducing the reinforced prediction mechanism with forecast weather information can significantly improve the ECPAS of prediction interval, but EAWAPI does not become significantly wider. Comparing the indicators of prediction intervals in different seasons, one can clearly see that the gains from the reinforced prediction mechanisms are more pronounced in the summer. In the summer of 2018-19, ECPAS is improved from 0.8850 (CTSGAN) to 0.9363 (RCTSGAN), and in the summer of 2019-20, ECPAS is improved from 0. 8916 to 0.9678. Thus, it can be suggested that in the case of inherently high electricity demand, volatility brought by weather factors is more likely to lead to volatility of electricity prices and occurrence of spikes, and by introducing reinforced prediction, prediction accuracy can be well improved.

## VI. Conclusion and further research

In this paper, a novel CTSGAN-based interval prediction method was proposed to generate realistic price scenarios. The probability density of each scenario between prediction intervals can be estimated. A novel reinforced prediction mechanism, in terms of volatility level of prediction weather factors information, was introduced to modify CTSGAN to improve the performance of prediction intervals. For some extremely high price spikes (>A\$350/MWh), the proposed RCTSGAN show high reliability. For normal electricity prices, the proposed RCTSGAN or CTSGAN shows low sharpness. Numerical simulation demonstrates that for electricity prices of different seasons, the proposed method shows superiority. In particular, the RCTSGAN method brings a more significant gain in predictive performance for summer electricity prices.